\documentclass[11pt, oneside]{article}   	
\usepackage[margin=0.8in]{geometry}                		
\usepackage{graphicx}				
\usepackage{subcaption}
\usepackage[font=footnotesize]{caption}
\usepackage{float}
\usepackage{amssymb}
\usepackage{hyperref}
\hypersetup{
    colorlinks=true,
    linkcolor=blue,
    filecolor=blue,      
    urlcolor=blue,
}

\title{\textbf{Defense Against Adversarial Attacks using \\Convolutional Auto-Encoders}}

\author{
Shreyasi Mandal\\
{(Indian Institute of Technology, Kanpur)}
}

\begin{document}

\maketitle

\abstract{Deep learning models, while achieving state-of-the-art performance on many tasks, are susceptible to adversarial attacks that exploit inherent vulnerabilities in their architectures. Adversarial attacks manipulate the input data with imperceptible perturbations, causing the model to misclassify the data or produce erroneous outputs. This work is based on enhancing the robustness of targeted classifier models against adversarial attacks. To achieve this, an convolutional autoencoder-based approach is employed that effectively counters adversarial perturbations introduced to the input images. By generating images closely resembling the input images, the proposed methodology aims to restore the model's accuracy.}
    
\section{Introduction}
\label{sec:intro}
\begin{figure}[!ht]
\centering
\captionsetup{justification=centering}
  \includegraphics[width=9cm]{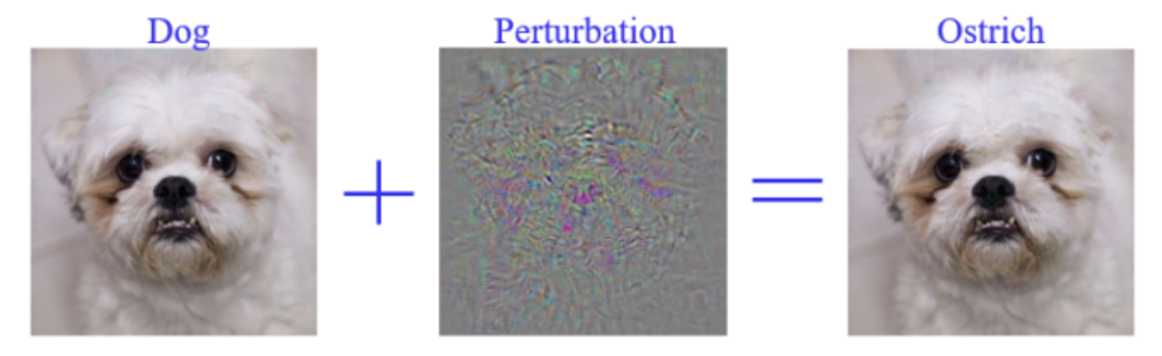}
  \caption{Szegedy et al. \cite{szegedy2013intriguing} were able to fool AlexNet \cite{krizhevsky2012imagenet}\\ by classifying a perturbed image of a dog into an ostrich}
  \label{fig:fool_eg}
\end{figure}

Deep learning has made notable progress across a diverse range of machine learning domains, including image classification, object detection and speech recognition. The influence of deep learning extends to an increasing number of real-world applications and systems. However, recent research highlights the susceptibility of deep learning models to well-designed input samples, called adversarial examples. Adversarial examples are imperceptible to humans but can easily fool deep neural networks \cite{yuan2019adversarial}.
\\
Szedegy et al. \cite{szegedy2013intriguing} manipulated state-of-the-art deep neural networks to misclassify images by applying subtle and barely perceptible perturbations, as seen in Figure \ref{fig:fool_eg}, showcasing the vulnerability of these models to targeted errors. Researchers discovered that a single perturbation could lead a neural network to misclassify multiple classes, revealing inherent weaknesses in the training algorithms. Despite neural networks being nonlinear functions \cite{grossberg1988nonlinear}, it was surprising to find that linear approximations are effective in creating adversarial perturbations. This unexpected result is attributed to the design of easily trainable models, which empirically exhibit an accurate local linear approximation for their loss function \cite{goodfellow2014explaining}, allowing adversaries to make subtle changes in the pixel values that lead to misclassification.
\\
There has been a good amount of progress in research pertaining to defending against these adversarial attacks. The defenses aim to either detect whether an input is adversarial, or aim to modify the input so that it is no longer adversarial in nature. Adversarial attackers on image classifiers fool neural networks by perturbing inputs in the direction of the gradient of the target model \cite{graves2020image}. As a result, the perturbed image is a data-point that maps to a different distribution than that of the original image dataset.
\\
Adversarial attacks pose a significant threat to the security and reliability of machine learning systems, particularly in domains such as autonomous vehicles, fraud detection, and cybersecurity. As such, researchers and practitioners in the field of machine learning are actively working to develop defenses against these attacks. In this paper, two kinds of adversarial attacks namely Fast Gradient Sign Method (FGSM) \cite{goodfellow2014explaining} and Projected Gradient Descent (PGD) \cite{madry2017towards} have been explored on the MNIST \cite{mnist} and Fashion-MNIST\cite{xiao2017fashion} dataset and are defended using convolutional auto-encoders \cite{bank2023autoencoders}.

\section{Related Work}
\label{sec:rw}

\subsection{Adversarial Attacks}
Akhtar et.al. \cite{akhtar2018threat} proposes a survey of adversarial attacks on deep learning in computer vision such as Box Constrained L-BFGS \cite{fletcher2000practical}, FGSM \cite{goodfellow2014explaining}, Jacobian based Saliency Map Attack \cite{papernot2016limitations}, One-Pixel Attack \cite{su2019one} and Carlini and Wagner Attacks \cite{carlini2017towards}. Despite the high accuracy of deep neural networks, they can be vulnerable to input perturbations that can cause significant changes in their outputs. The article reviews recent contributions that devise attacks and defenses for deep learning, particularly in safety and security-critical applications.
\\
There are two main types of adversarial attacks: targeted \cite{li2020towards} and untargeted \cite{wu2019untargeted}, and two main categories of attacks based on the attacker's knowledge of the model: white-box \cite{ebrahimi2017hotflip} and black-box attacks \cite{ilyas2018black}. In a targeted attack, the attacker tries to manipulate the model's output so that it produces a specific, predefined result. In contrast, in an untargeted attack, the attacker aims to simply cause the model to make a mistake, without specifying what that mistake should be. A white-box attack occurs when the attacker has complete knowledge of the model, including its architecture, parameters, and training data. This information can be used to craft a specific attack that exploits weaknesses in the model. In contrast, a black-box attack is one in which the attacker has little or no knowledge of the model's internals, and must rely on input/output observations to craft an attack.
\\
The Fast Gradient Sign Method (FGSM) is a gradient-based technique designed for efficient computation of norm-bounded perturbations, prioritizing computation efficiency over achieving high fooling rates. Goodfellow et al. \cite{goodfellow2014explaining} utilized FGSM to support their linearity hypothesis, suggesting that the linear behavior induced by components like ReLUs \cite{agarap2018deep} in high-dimensional spaces makes modern neural networks vulnerable to adversarial perturbations. Fast Gradient Sign Method computes $\rho$ as : 
$$\rho = \epsilon \: sign(\nabla \mathbb{J} (\theta, I_c, l))$$
where gradient of cost function around the current value of model parameters  w.r.t. $I_c$ and $\epsilon$ bounds the $l_\infty$ norm. Contrary to the prevailing notion of non-linearity causing vulnerability, FGSM has become a key influencer in white-box attack scenarios due to its foundational concept of gradient ascent on the model's loss surface for deception. 
\\
An intuitive extension of FGSM (single large step in the direction to increase loss) is to iteratively take multiple small steps while adjusting the direction after each step. The Basic Iterative Method (BIM) \cite{kurakin2018adversarial} does exactly that, and iteratively computes the following: 
$$I_{\rho}^{i+1} = Clip_{\epsilon} \{I_{\rho}^i + \alpha \; sign(\nabla \mathbb{J} (\theta, I_{\rho}^i, l))\} $$
where $I_{\rho}^i$ denotes the perturbed image at the $i^{th}$ iteration. $Clip_{\epsilon}\{.\}$ clips the values of the pixels of the image in its argument at $\epsilon$ and $\alpha$ determines the step size (normally $\alpha = 1$). 
\\
The PGD adversarial attack was originally proposed by Madry et al. \cite{madry2017towards} as the most strongest “first-order adversarial” attack. Actually, PGD is a well-known optimization technique projecting gradients in a ball. PGD is basically the same as BIM, described above, except that PGD starts at a random point in the ball and perform random restarts. The 'l' parameter in the above equation is emphasized, because adversarial attacks can trick the ML classifier either as targeted or untargeted. In the targeted case, the malicious attacker forces the classifier to produce a certain class of output. On the other hand, the classifier is expected to give a misclassification where the label of the output class is not important in the untargeted case.
\\
A relaxation term is introduced to the standard loss in Projected gradient Descent (PGD) and \textbf{FGSM} which finds more suitable gradient-directions, increases attack efficacy and leads to more efficient adversarial training.
\\
Considering $L_2$ norm distortions, the Carlini and Wagner \cite{carlini2017towards} attack is presently the most effective white-box attack in the literature. However, this method is slow since it performs a line-search for one of the optimization terms, and often requires thousands of iterations. Rony et.al. \cite{ddn} proposes an efficient approach to generate gradient-based attacks that induce misclassifications with low $L_2$ norm, by decoupling the direction and the norm of the adversarial perturbation that is added to the image. GeoDA \cite{geoda} is a geometric framework to generate adversarial examples in one of the most challenging black-box settings where the adversary can only generate a small number of queries, each of them returning the top-1 label of the classifier.

\subsection{Adversarial Defense}
Deep learning algorithms have been shown to perform extremely well on many classical machine learning problems. However, recent studies have shown that deep learning, like other machine learning techniques, is vulnerable to adversarial samples: inputs crafted to force a deep neural network (DNN) to provide adversary-selected outputs. Such attacks can seriously undermine the security of the system supported by the DNN, sometimes with devastating consequences \cite{apruzzese2019addressing}. For example, autonomous vehicles can be crashed, illicit or illegal content can bypass content filters, or biometric authentication systems can be manipulated to allow improper access. Akhtar and Mian \cite{akhtar2018threat} categorized defenses against adversarial attacks into three main groups. These include (1) modifying target models for robustness, (2) altering inputs to remove perturbations, and (3) integrating external modules, primarily detectors, into the model. Since 2018, the trajectory of adversarial defense research has predominantly followed these three avenues of approach.\\
Papernot et.al. \cite{defense-dist} introduces a defensive mechanism called defensive distillation to reduce the effectiveness of adversarial samples on DNNs. It reduces the amplitude of the network gradient exploited by adversaries to craft adversarial samples. Distillation Temperature (T) hyperparameter is introduced in the softmax layer to reduce the amplitude of the gradients.\\
PuVAE \cite{puvae} uses Variational Autoencoder \cite{doersch2016tutorial} to remove the perturbation from Adversarial Examples and then classify the image. During training, the distribution of latent space is learnt for each class. At inference time, a latent vector z is sampled for every class, which is passed through the decoder and subsequently, the encoder. The class label is then selected corresponding to the closest projection. 
$$ y* = argmin_{y_i \in C} D(x, x'_{y_i})$$
The y* is termed as a ‘purified’ sample which is then sent to the image classifier.
\\
Defense GANS \cite{defense-gan} use the generator part of GAN to generate an image very close to the input (adversarial) image, and then send it to the classification model. The model is trained on the real images using Wasserstein loss (WGAN) \cite{frogner2015learning} instead of the usual Jensen-Shannon Divergence (JSD) \cite{menendez1997jensen}. At the inference time, prior to feeding an image x to the classifier, it is projected onto the range of generator by minimizing the reconstruction error $min_z ||G(z) - x||_2^2$. The reconstructed image is then sent to the classifier.

\section{Methodology}
\label{sec:meth}
\subsection{Dataset}
 The publicly available pre-processed MNIST database \cite{mnist} of handwritten digits and the Fashion-MNIST database \cite{xiao2017fashion} of Zalando's article images is used to implement the proposed methodology. 
 \\
 The MNIST database of handwritten digits, has a training set of 60,000 examples, and a test set of 10,000 examples. It is a subset of a larger set available from NIST. The digits have been size-normalized and centered in a fixed-size image. Each example is a 28x28 grayscale image, associated with a label from 10 classes.
 \\
The Fashion-MNIST database is designed to seamlessly replace the original MNIST dataset in benchmarking machine learning algorithms. It mirrors the same image size, structure, and training/testing splits, facilitating a straightforward integration for comparative analysis.

\begin{figure}[!ht]
\centering
  \begin{subfigure}[b]{0.25\textwidth}
    \includegraphics[width=\textwidth]{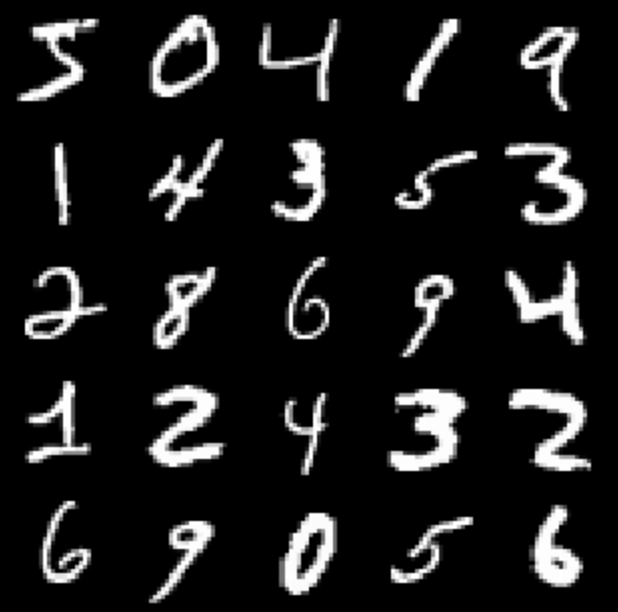}
    \caption{}
    \label{fig:1}
  \end{subfigure}
  \qquad
  \begin{subfigure}[b]{0.25\textwidth}
    \includegraphics[width=\textwidth]{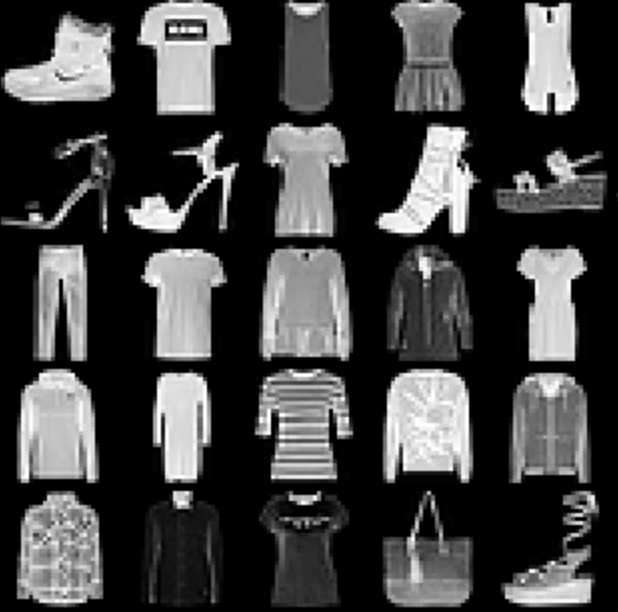}
    \caption{}
    \label{fig:2}
  \end{subfigure}
  \caption{Publicly available datasets (a) MNIST and (b) Fashion MNIST}
    \label{fig:dataset}
\end{figure}

\subsection{Effect of Attacks:} 
To evaluate the robustness of the proposed methodology, it is tested against two types of adversarial attacks: the Fast Gradient Sign Method (FGSM) \cite{goodfellow2014explaining} and Projected Gradient Descent (PGD) \cite{madry2017towards}. Both FGSM and PGD attacks are white box untargeted attacks.
\\
For training the databases, VGG-16 \cite{simonyan2014very} is used for multi-class classification. Transfer learning \cite{torrey2010transfer} is used to get an accuracy of 99\% on the test dataset for MNIST and an accuracy of 92\% on the test dataset for Fashion-MNIST.

\subsubsection{FGSM Attack}
Table \ref{tab:fgsm} below show the effect of epsilon ($\epsilon$) on the accuracy of the MNIST and the Fashion-MNIST dataset with FGSM attack. As we can see, the accuracy decreases to \textbf{5.23\%} from \textbf{99.01\%} as the value of $\epsilon$ increases to \textbf{1.50} from 0 for the MNIST dataset. For the Fashion-MNIST dataset, the accuracy decreases to \textbf{4.12\%} from \textbf{91.22\%} as the value of $\epsilon$ increases to \textbf{1.50} from 0. 
\begin{figure}[!ht]
\centering
\captionsetup{justification=centering}
\begin{subfigure}[b]{0.8\textwidth}
   \includegraphics[width=1\linewidth]{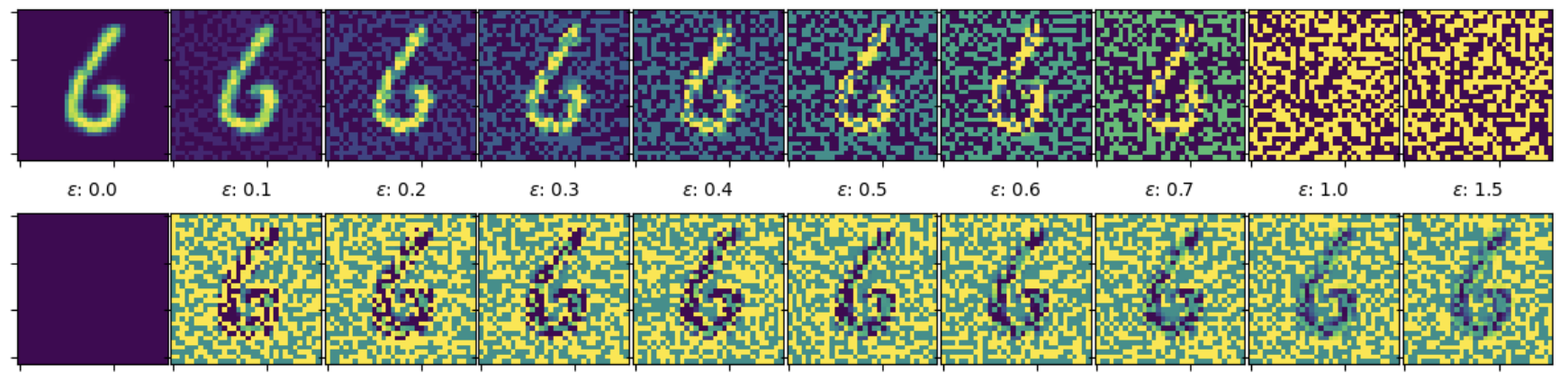}
   \caption{}
   \label{fig:fgsm-m} 
\end{subfigure}

\begin{subfigure}[b]{0.8\textwidth}
   \includegraphics[width=1\linewidth]{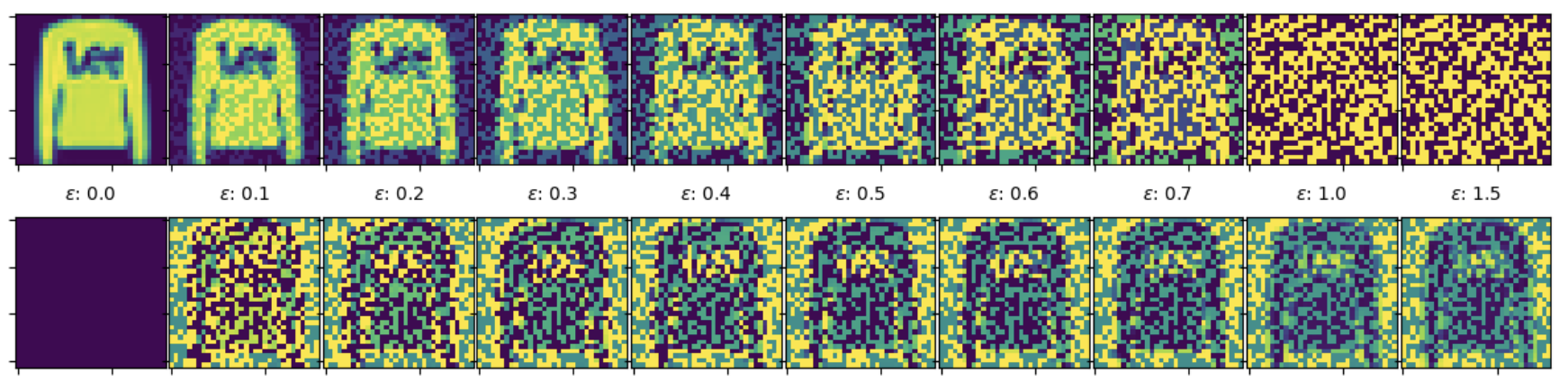}
   \caption{}
   \label{fig:fgsm-fm}
\end{subfigure}
\caption{FGSM Attack with different $\epsilon$ values on (a) MNIST dataset (b) Fashion-MNIST dataset. \\The images in the upper row denote the adversarial images, while those on the lower row correspond to \\the perturbation added to the original image. The top left image corresponds to the original image ($\epsilon = 0$).}
\label{fig:attack}
\end{figure}
\\
As evident from Fig \ref{fig:attack}, using large epsilon ($\epsilon$) values leads to corruption of the label semantics making it impossible to retrieve the original image. Hence, the $\epsilon$ value used in the workflow is 0.60. 

\begin{table}[!ht]
    \begin{subtable}[!ht]{0.5\textwidth}
        \centering
        \begin{tabular}{|c|c|c|}
        \hline
        \multicolumn{3}{|c|}{$\epsilon$ vs Accuracy for FGSM attack} \\
        \hline
        \textbf{Epsilon} & \multicolumn{2}{|c|}{\textbf{Accuracy} (w/o defense)}  \\
        \cline{2-3}
         & MNIST & Fashion-MNIST \\
        \hline
        0.00 & 0.9901 & 0.9122 \\
        0.10 & 0.9099 & 0.4751 \\
        0.20 & 0.8237 & 0.3798\\
        0.30 & 0.7274 & 0.2973\\
        0.40 & 0.5840 & 0.2376\\
        0.50 & 0.4054 & 0.1889\\
        0.60 & 0.2648 & 0.1417\\
        0.70 & 0.1610 & 0.0955\\
        1.00 & 0.0531 & 0.0444\\
        1.50 & 0.0523 & 0.0412\\
        \hline
        \end{tabular}
        \caption{}
        \label{tab:fgsm}
    \end{subtable}
    \hspace{-1cm}
    \begin{subtable}[!ht]{0.5\textwidth}
        \centering
        \begin{tabular}{|c|c|c|}
        \hline
        \multicolumn{3}{|c|}{$\epsilon$ vs Accuracy for PGD attack} \\
        \hline
        \textbf{Epsilon} & \multicolumn{2}{|c|}{\textbf{Accuracy} (w/o defense)}  \\
        \cline{2-3}
         & MNIST & Fashion-MNIST \\
        \hline
        0.00 & 0.9922 & 0.9206 \\
        0.05 & 0.6130 & 0.3150 \\
        0.10 & 0.1723 & 0.2962\\
        0.15 & 0.0418 & 0.2942\\
        0.20 & 0.0181 & 0.2940\\
        0.30 & 0.0118 & 0.2934\\
        \hline
        \end{tabular}
        \caption{}
        \label{tab:pgd}
    \end{subtable}
     \caption{(a) FGSM Attack and (b) PGD Attack on MNIST and Fashion-MNIST datasets}
     \label{tab:Attack}
\end{table}


\subsubsection{PGD Attack}
PGD is akin to i-FGSM i.e “many mini FGSM steps”. A slight difference lies in the optimization algorithm used in this approach, compared to FGSM. FGSM uses normal gradient descent steps. PGD in contrast, runs projected (normalized) steepest descent under the $l_\infty$ norm.
\\
Table \ref{tab:pgd} below show the effect of epsilon ($\epsilon$) on the accuracy of the MNIST and the Fashion-MNIST dataset with PGD attack. As we can see, the accuracy decreases to \textbf{1.18\%} from \textbf{99.22\%} as the value of $\epsilon$ increases to \textbf{0.30} from 0 for the MNIST dataset. For the Fashion-MNIST dataset, the accuracy decreases to \textbf{29.34\%} from \textbf{92.06\%} as the value of $\epsilon$ increases to \textbf{0.30} from 0. 

\begin{figure}[!ht]
\centering
\captionsetup{justification=centering}
\begin{subfigure}[b]{0.63\textwidth}
   \includegraphics[width=1\linewidth]{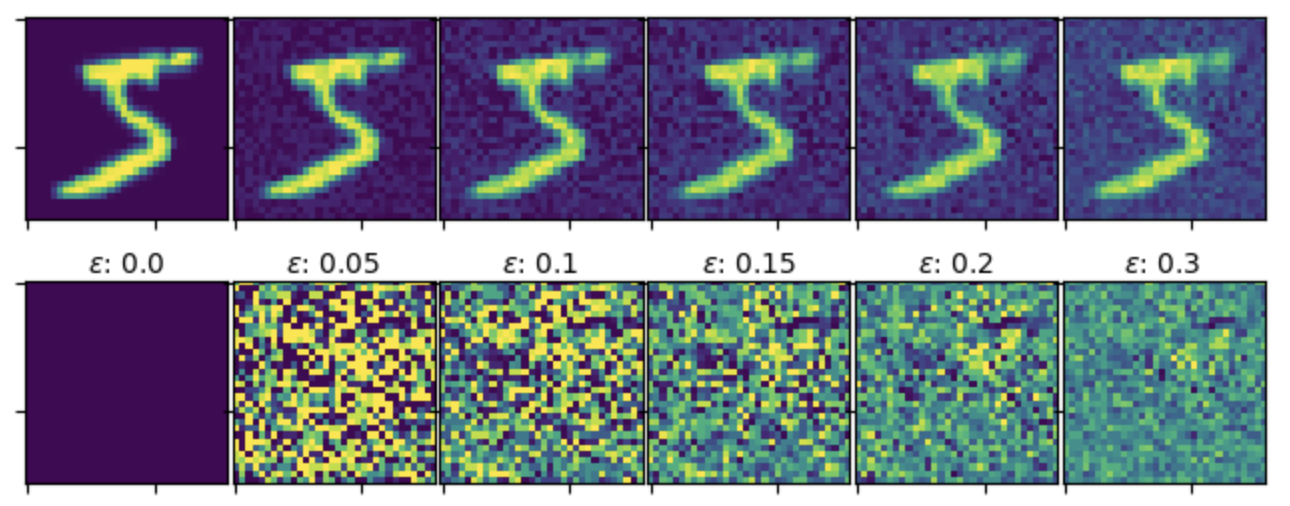}
   \caption{}
   \label{fig:fgsm-m} 
\end{subfigure}

\begin{subfigure}[b]{0.63\textwidth}
   \includegraphics[width=1\linewidth]{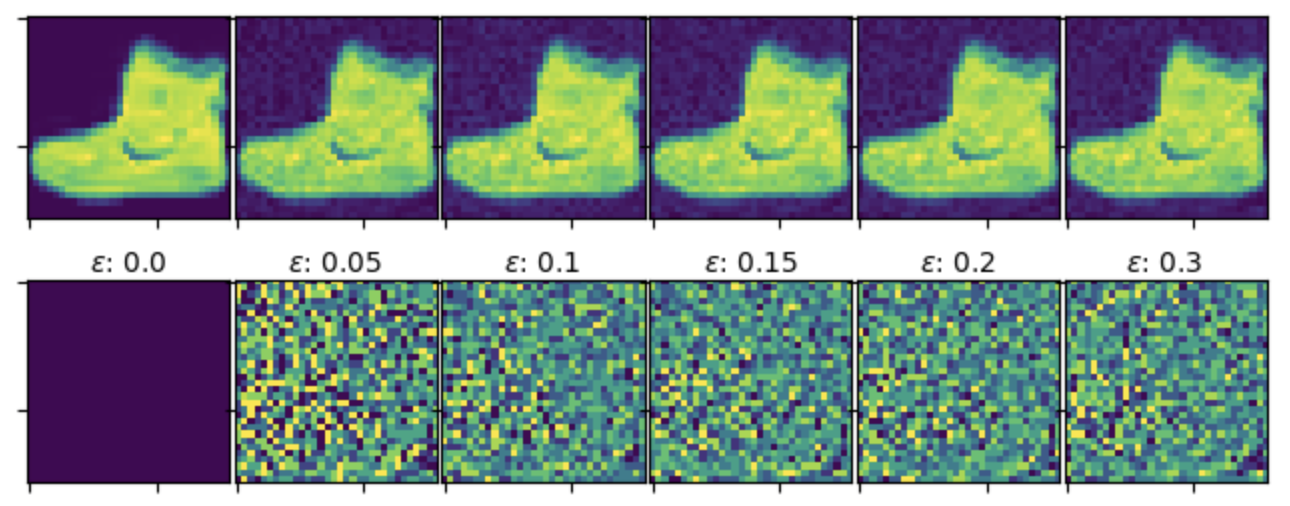}
   \caption{}
   \label{fig:fgsm-fm}
\end{subfigure}
\caption{PGD Attack with different $\epsilon$ values on (a) MNIST dataset (b) Fashion-MNIST dataset. \\The images in the upper row denote the adversarial images, while those on the lower row correspond to \\the perturbation added to the original image. The top left image corresponds to the original image ($\epsilon = 0$).}
\label{fig:attack}
\end{figure}

\subsection{Proposed Defense Architecture}
\begin{figure}[!ht]
\centering
\captionsetup{justification=centering}
  \includegraphics[width=15cm]{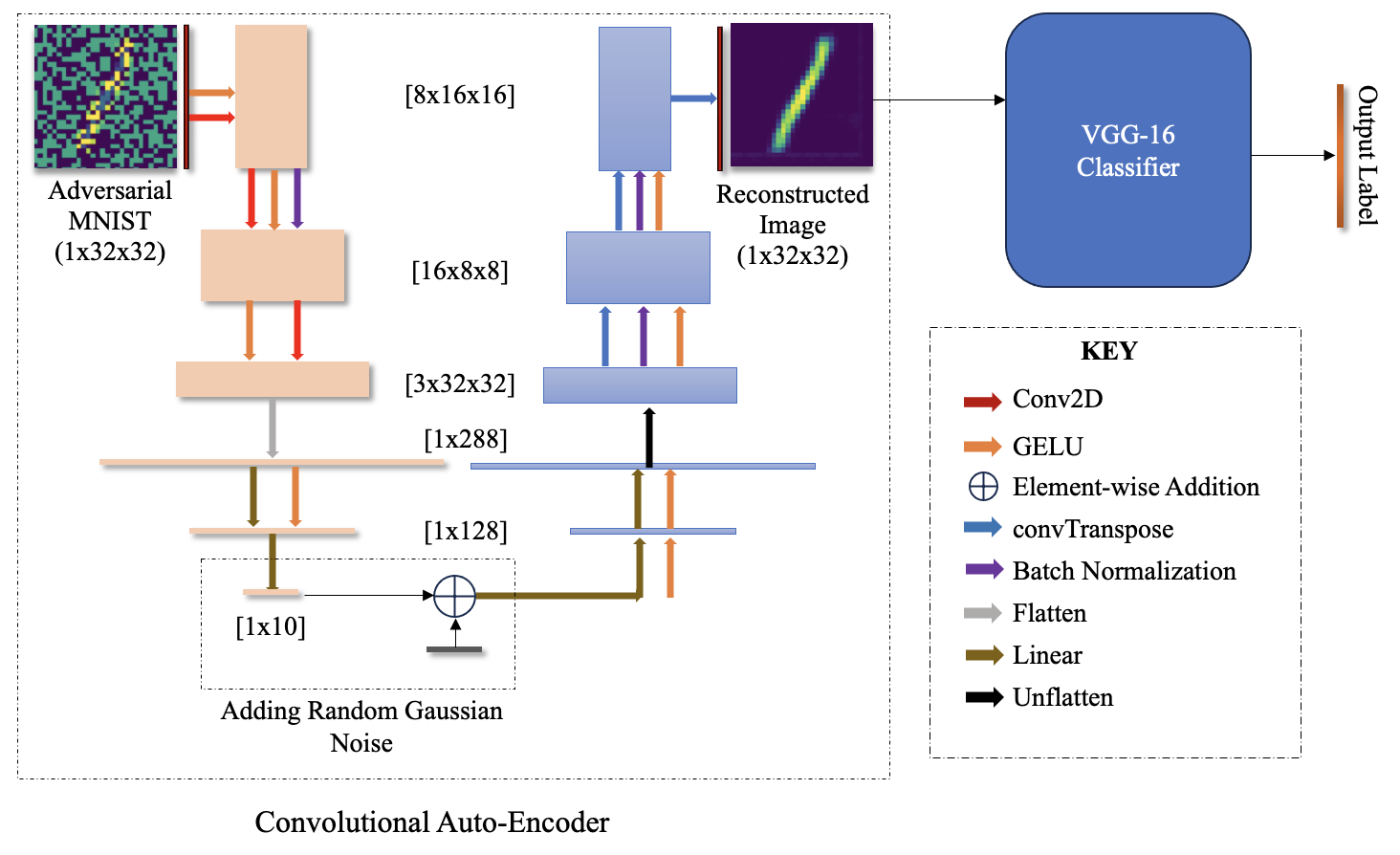}
  \caption{Proposed Defense framework with a U-shaped Convolutional Auto-Encoder and \\a pre-trained VGG-16 Classifier}
  \label{fig:archi}
\end{figure}

\begin{figure}[!ht]
\centering
\captionsetup{justification=centering}
\begin{subfigure}[b]{0.9\textwidth}
   \centering
   \begin{subfigure}[b]{0.45\textwidth}
    \includegraphics[width=\textwidth]{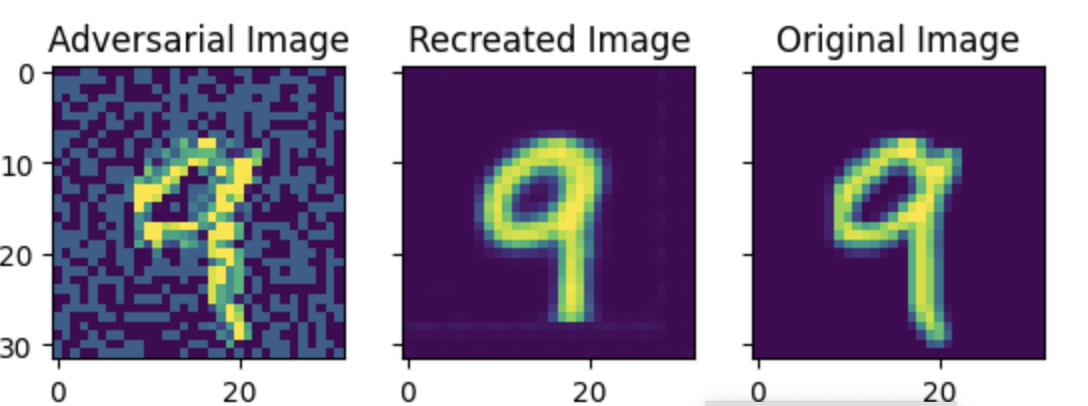}
    \caption{}
    \label{fig:1}
  \end{subfigure}
  \begin{subfigure}[b]{0.45\textwidth}
    \includegraphics[width=\textwidth]{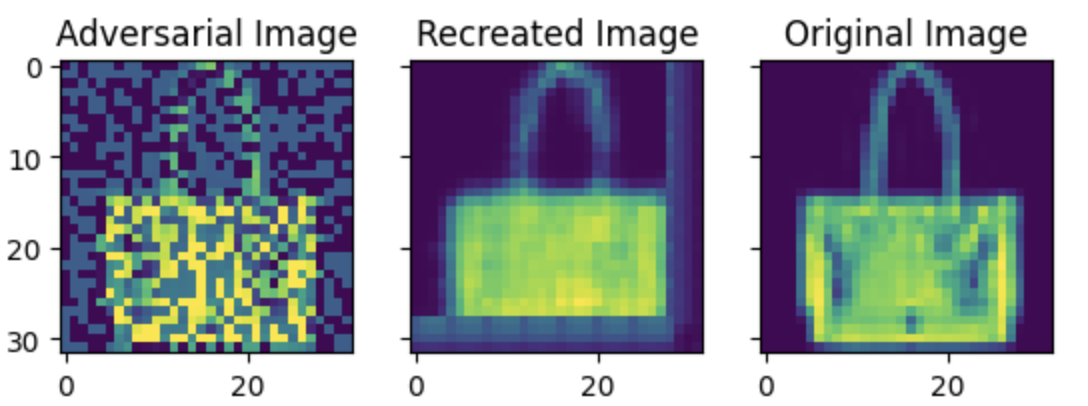}
    \caption{}
    \label{fig:2}
  \end{subfigure}
\end{subfigure}

\begin{subfigure}[b]{0.9\textwidth}
   \centering
    \begin{subfigure}[b]{0.45\textwidth}
    \includegraphics[width=\textwidth]{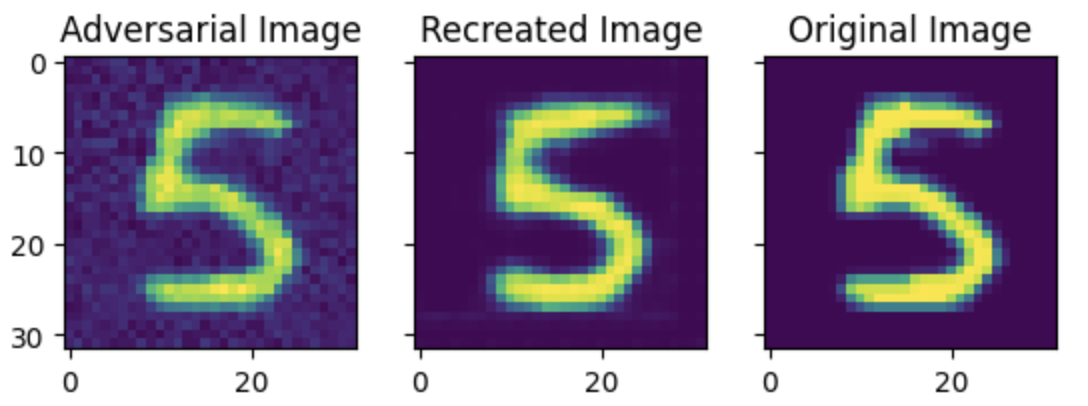}
    \caption{}
    \label{fig:1}
  \end{subfigure}
  \begin{subfigure}[b]{0.45\textwidth}
    \includegraphics[width=\textwidth]{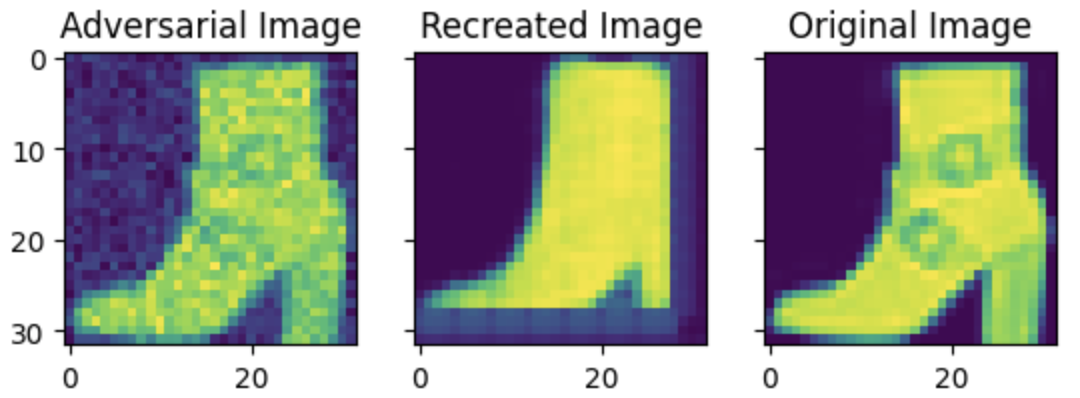}
    \caption{}
    \label{fig:2}
  \end{subfigure}
\end{subfigure}
\caption{Reconstructed images from the proposed auto-encoder architecture during inference time. \\(a) and (b) shows the recreated images from FGSM attacked MNIST and Fashion-MNIST images \\respectively for $\epsilon = 0.60$. (c) and (d) shows the recreated images from PGD attacked MNIST\\ and Fashion-MNIST images respectively for $\epsilon = 0.15$}
\label{fig:reconstruction}
\end{figure}

An U-shaped convolutional auto-encoder as shown in Fig \ref{fig:archi} is used to reconstruct original input from the adversarial image, effectively removing the adversarial perturbations. The goal of the autoencoder network is to minimise the mean squared error loss \cite{error2010mean} between the original unperturbed image and the reconstructed image, which is generated using an adversarial example. While doing so, a random Gaussian noise \cite{guo2011estimation} is added after encoding the image so as to make the model more robust. The idea behind adding the noise is to perturb the latent representation by a small magnitude, and then decode the perturbed latent representation, akin to how the adversarial examples are generated in the first place.
\\
During inference, the adversarial image generated from the original image is fed to the trained auto-encoder to get the reconstructed image close to the original image as evident from Fig \ref{fig:reconstruction}. This reconstructed images are then sent to the pre-trained VGG-16 classifier network.
\\
The GELU \cite{hendrycks2020gaussian} activation layers are used in the model architecture. GELU being a smoother activation function having a non-zero derivative for all inputs helps to address the dying ReLU problem and allow for more effective learning in deep neural networks.

\section{Results}
\label{sec:result}

Table \ref{tab:defense} shows the model's performance on the MNIST and Fashion-MNIST dataset. The accuracy of the pre-trained VGG-16 classifier on the MNIST and Fashion-MNIST dataset with FGSM attack increases by 65.61\% and 59.76\% respectively. For the PGD attack, the accuracy increases by 89.88\% and 43.49\%. This shows the efficacy of our model in defending the adversarial attacks with high accuracy.
\begin{table}[!ht]
\centering
\begin{tabular}{|c|c|c|c|c|}
\hline
\textbf{Attack} & \multicolumn{2}{|c|}{\textbf{Accuracy} (w/o defense)} & \multicolumn{2}{|c|}{\textbf{Accuracy} (with defense)}  \\
\cline{2-5}
 & MNIST & Fashion-MNIST  & MNIST & Fashion-MNIST \\
\hline
FGSM ($\epsilon=0.60$) & 0.2648 & 0.1417 & 0.9209 & 0.7393\\
PGD ($\epsilon=0.15$) & 0.0418 & 0.2942 & 0.9406 & 0.7291\\
\hline
\end{tabular}
\caption{Accuracy of VGG-16 on MNIST and Fashion-MNIST dataset}
\label{tab:defense}
\end{table}
\\
Table \ref{tab:comp} shows the comparison of the proposed convolutional auto-encoder architecture with the existing state-of-the-art models, mainly Defense GANS \cite{defense-gan} and PuVAE \cite{puvae} for FGSM attack with $\epsilon=0.60$ on the MNIST dataset.  As evident from the table, the proposed architecture has more accuracy and a much lower latency compared to the other models, particularly in applications like autonomous driving where quick and precise decisions are crucial.

\begin{table}[!ht]
\centering
\begin{tabular}{|c|c|c|c|c|}
\hline
Criteria & No Attack & Defense GAN & PuVAE & Our Model \\
\hline
Accuracy  & 0.9901 & 0.8529 & 0.8133 & \textbf{0.9209} \\
\hline
Inference Latency  & - & 14.80 s & 0.11 s & \textbf{0.008 s} \\
\hline
\end{tabular}
\caption{Comparison of the proposed architecture with existing SOTA models}
\label{tab:comp}
\end{table}

\section{Conclusion and Future Work}
The employed convolutional autoencoder-based approach effectively counters adversarial perturbations, restoring the model's accuracy notably. With the FGSM attack, accuracy on MNIST and Fashion-MNIST datasets increases by 65.61\% and 59.76\% respectively, and with the PGD attack, by 89.88\% and 43.49\%. This signifies the model's robust defense against adversarial attacks, showcasing its high accuracy. 
\\
However, unlike Defense GANs, which adeptly handle various attack types during inference, our model operates under the assumption of a specific attack type, like FGSM with $\epsilon = 0.60$. This difference in approach might limit our model's adaptability to diverse adversarial scenarios encountered in practical settings. Moreover, models like PuVAEs, with their increased complexity, tend to achieve higher robustness, especially when dealing with intricate datasets.
\\
Potential future work might include incorporating bayesian machine learning techniques in the proposed architecture to make it more robust to attacks, while keeping the inference latency in check as well.

\newpage

\label{sec:ref}
\bibliography{references} 
\bibliographystyle{ieeetr}

\end{document}